\documentclass{article} 
\usepackage{iclr2027_conference,times}

\usepackage{amsmath,amsfonts,bm}

\def\eqref#1{equation~\ref{#1}}

\def\1{\bm{1}}

\DeclareMathAlphabet{\mathsfit}{\encodingdefault}{\sfdefault}{m}{sl}
\SetMathAlphabet{\mathsfit}{bold}{\encodingdefault}{\sfdefault}{bx}{n}

\usepackage{tcolorbox}
\tcbuselibrary{skins,breakable}
\usepackage{listings}
\usepackage{xcolor}

\newtcolorbox{promptbox}[1]{
    breakable,
    skin=enhanced,
    colback=white,
    colframe=black,
    coltitle=white,
    colbacktitle=black,
    title=#1,
    fonttitle=\bfseries,
    left=2pt,right=2pt,top=4pt,bottom=4pt,
    width=\linewidth
}

\usepackage{amssymb}
\DeclareMathOperator{\LeakyReLU}{LeakyReLU}
\usepackage{wrapfig}
\usepackage{graphicx}
\usepackage[utf8]{inputenc} 
\usepackage[T1]{fontenc}    
\usepackage{hyperref}       
\usepackage{url}            
\usepackage{booktabs}       
\usepackage{amsfonts}       
\usepackage{amsmath}
\usepackage{nicefrac}       
\usepackage{microtype}      
\usepackage{xcolor} 
\usepackage{soul}
\usepackage{caption}
\usepackage{wrapfig}
\usepackage{makecell}
\usepackage{longtable}
\usepackage{multirow}
\usepackage{subfig}
\usepackage{threeparttable}
\usepackage[ruled,vlined]{algorithm2e}
\definecolor{commentgray}{RGB}{0,128,0}
\definecolor{mzy}{RGB}{255,187,255}

\SetCommentSty{mycommfont}
\SetKwComment{tcp}{// }{}
\usepackage[table]{xcolor}
\usepackage{array}
\usepackage{bm}

\providecommand{\bbobinlineval}[2]{#1$\pm$#2}
\providecommand{\bbobinlinebest}[2]{%
    \cellcolor{mzy}\textbf{#1}$\pm$#2}
\providecommand{\bbobinlinesecond}[2]{%
    \underline{#1}$\pm$#2}
\iclrfinalcopy
\title{Hyper Algorithm Design Agent: Evolving Learnable Optimizer from Zero}

\author{
\hspace{2mm} Zipei Yu$^{1}$, \quad Yue-Jiao Gong$^{1}$,\quad
Zeyuan Ma$^{2}$,\quad 
Yuncheng Jiang$^2$,\quad  
Zhiguang Cao$^{3}$\\
\quad \quad \quad $^1$ South China University of Technology\quad \quad
$^2$ South China Normal University \\
\quad \quad \quad \quad \quad \quad \quad \quad \quad$^3$ Singapore Management Univeristy\\
\quad \quad \texttt{\{zipei540, gongyuejiao\}@gmail.com}, \quad \texttt{mzy@ieee.org} \\ 
\quad \quad \quad \quad \texttt{ycjiang@scnu.edu.cn},\quad \texttt{zhiguangcao@outlook.com}
}

\begin{document}

\maketitle

\begin{abstract}
Meta-Black-Box Optimization (MetaBBO) is one of the highlights in the recent AI for Optimization trend. This paradigm's bi-level workflow leverages the learnable algorithm design policy at meta level to ensure the performance and generalization improvement on the low-level optimization task. While MetaBBO helps advance the performance lower bound of the resulted optimization system, it is currently handcrafted and customized case by case to adapt different optimization problems, which inevitably introduces inherent subjectivity and hence restricts the performance upper bound and usability in practice. In this paper, we address this issue  by regarding MetaBBO's design loop as coding task, where we could introduce openendedness into MetaBBO with recursive self-improvement capability of advanced coding agents. Specifically, we propose a dual-agent framework: i) a task agent continuously refines the codebase of a target MetaBBO approach through code evolution; ii) a hyper agent progressively modifies the task agent and itself to provide open-ended design behavior; iii) the evolved MetaBBO codebase is evaluated and all in-execution information is fed back to the agents for recursive self-referential improvement. As a result, given a naive MetaBBO template, our framework automates a design evolution and finds novel variants superior to up-to-date human-made MetaBBO baselines. Surprisingly, the experimental results also demonstrate that our framework supports fast adaption across different optimization domains. Solid interpretation analysis further reveals interesting design principles emerge in such open-ended process. This work serves as the first exploration on automating design of complex learning-assisted optimization algorithms.     
\end{abstract}

\section{Introduction}

Automated Algorithm Design~(AAD) has long been discussed in the optimization community~\citep{old-survey-3H,old-survey-thomas,old-survey-qiqi}, and has recently attracted growing attention with the emergence of novel paradigms such as Meta-Black-Box Optimization~(MetaBBO)~\citep{survey-our,survey-yang} and Large Language Model for Algorithm Design~(LLM4AD)~\citep{survey-niki,survey-liufei}. Despite diverse implementations, these paradigms share the same spirit: by introducing data-driven learning capabilities~(\emph{reinforcement learning}~\citep{rldas,gleet}, \emph{self-supervised learning}~\citep{glhf,okea}, \emph{in-context learning}~\citep{eoh,llamea}, etc.) into algorithm design, the resulting optimizers achieve robust performance gains and, more importantly, generalization across problems. This paper focuses on MetaBBO. MetaBBO adopts a bi-level learning-to-optimize architecture, where a neural network-based design policy~(e.g., a reinforcement learning agent~\citep{rl}) at the meta level dictates online design choices for the low-level optimizer. With performance-centric meta-learning over a problem distribution, the learned system appears to challenge the \emph{no-free-lunch}~(NFL) theorem~\citep{nfl-95,nfl-97}.


\begin{figure}[t]
    \centering   
    \includegraphics[width=0.98\textwidth]{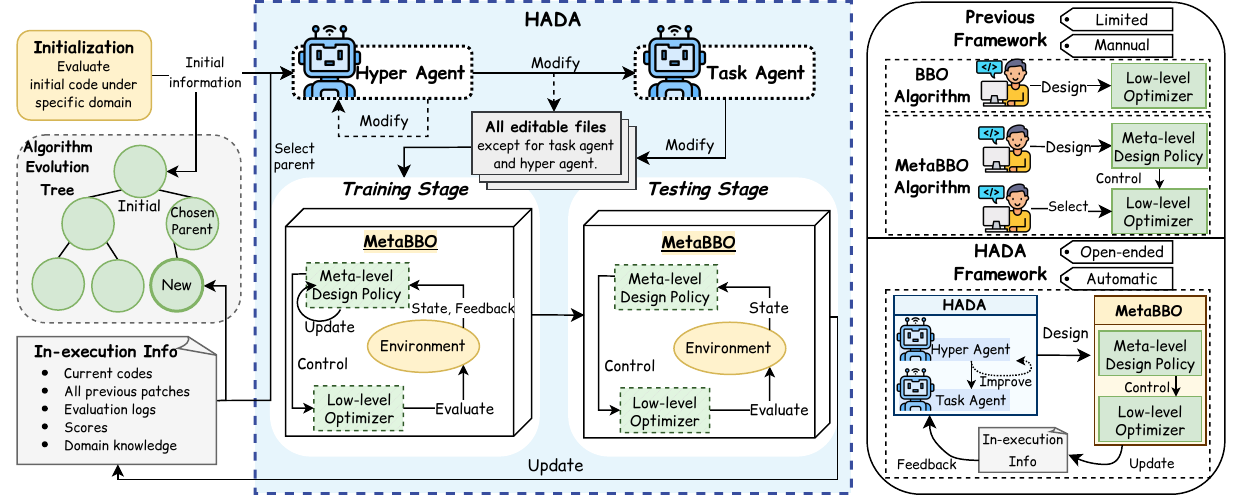}
    \caption{\textbf{Left}: The general workflow of our proposed HADA framework, which allows coordination between hyper agent and task agent for automatic MetaBBO design.
    \textbf{Right}: Essential differences between human-based design in BBO/MetaBBO and fully automated design in HADA.}
    \label{fig:intro}
    \vspace{-5mm}
\end{figure}

\textbf{However, is this true?} Admittedly, MetaBBO reduces human reliance compared to traditional expert-driven design~(e.g., dynamic algorithm configuration~\citep{madac,metabbo-ac-1}, algorithm selection~\citep{metabbo-as-1,metabbo-as-2}, and algorithm generation~\citep{designx}). Yet, as shown in the right part of Fig.~\ref{fig:intro}, the MetaBBO system itself still depends heavily on human experience: selecting the low-level optimizer requires expertise in algorithm-problem performance analysis, and the meta-level pipeline introduces further hand-crafted designs~(e.g., decision features, policy networks). \textbf{Can we really resolve the NFL issue in MetaBBO?} It motivates our work.

Our solution is to introduce openendedness~\citep{open-schmidhuber}---algorithmic systems pursuing never-ending innovation~\citep{open-stanley-1,open-stanley-2}---into MetaBBO. At its core lies a self-referential, continual self-improvement mechanism, exemplified by G\"odel machines~\citep{godel-1} and later extended to self-reflective interpreters~\citep{godel-2} and LLM-assisted generative approaches~\citep{open-DGM,open-hyperagents}.

Following this lead, we propose \textbf{HADA} (\textbf{H}yper \textbf{A}lgorithm \textbf{D}esign \textbf{A}gent), the first framework to reduce MetaBBO's reliance on human expertise. As shown in Fig.~\ref{fig:intro}, HADA runs a search loop over the MetaBBO algorithm space. This poses two main challenges: how to represent the vast algorithmic space, and how to achieve open-ended and effective search. To address the first, we treat each MetaBBO algorithm as a sourcecodes project, reducing algorithm design to coding tasks that advanced coding agents~\citep{qwen,deepsek-v3,gemini} can readily operate on. For the second, we propose a closed-loop search paradigm based on a bi-agent system: a Task Agent designs better MetaBBO algorithms by editing the corresponding sourcecodes, while a Hyper Agent controls the Task Agent's reasoning logic by editing Task Agent's prompt file, and moreover, itself. This yields an open-ended, self-improving algorithm evolution process. Additionally, a tree-based editing history balances exploration and exploitation via weighted parent sampling, and unified in-execution information makes HADA domain-agnostic. With these designs, HADA autonomously discovers novel MetaBBO algorithms. We summarize our major contributions as follows:

\begin{itemize}
    \item \textbf{Paradigm Innovation}: As shown in the right side of Fig.~\ref{fig:intro}, HADA introduces significant paradigm shift: the bi-agent open-ended self-improvement paradigm prevents MetaBBO from being stuck with labor-intensive meta/lower level design.  
    \item \textbf{Coherent Methodology}: We have carefully designed each part in HADA. This includes but not limited to the proposed open-ended bi-agent union, the code level recursive evolution pipeline, the tree-based history maintenance etc. The overall methodology holds both simplicity and exceptional capability.
    \item \textbf{Performance Breakthrough}: Given a naive MetaBBO baseline~(a ``zero'' one) as the start point, HADA successfully evolved high-performance MetaBBO variants for three optimization domain: single-objective, constrained, multi-objective problems, achieving 121.7\%, 114.7\% and 104.3\% performance leap compared to existing state-of-the-art baselines respectively. With the openendedness, HADA also supports cold-start adaption either for cross-domain scenario or from off-the-shelf baselines in practice.  
    \item \textbf{Clear Reproducibility}: We opensource HADA's codebase, the sourcecodes and neural parameters of the discovered MetaBBO algorithms at \textcolor{orange}{\url{https://github.com/MetaEvo/HADA-AAD}},  for the ease of future users.
\end{itemize}

\section{Related Works}

\subsection{Meta-Black-Box Optimization}\label{sec:2.1}
The ambition of AAD predates modern machine learning. Early works---\emph{algorithm selection}~\citep{rice1976algorithm}, \emph{hyper-heuristics}~\citep{burke2013hyperheuristics}, \emph{Programming by Optimisation}~(PbO)~\citep{hoos2012programming}, and \emph{automated algorithm configuration}~\citep{hutter2019automated}---share the AAD premise: algorithms are objects to be searched over rather than hand-crafted artifacts. However, they operate over fixed design spaces and target only a single (or a few) problem instance(s). MetaBBO~\citep{survey-our,survey-yang} is a recent AAD avenue whose key advantage is generalization: it integrates a training problem distribution $\mathcal{P}$ into a bi-level learning-assisted optimization architecture. With the meta-level design policy parameterized as $\pi_\theta$, the meta-learning objective is formulated as:
\begin{equation}\label{eq:metabbo}
    \mathcal{J}(\theta) = \mathbb{E}_{p\in \mathcal{P}} \sum_{t=1}^{T} \mathcal{R}(s_t,\pi(s_t;\theta)|p),
\end{equation}
where $T$ is the optimization horizon, $s_t$ is the state abstracted from the $t$-th optimization step of instance $p$, and $\mathcal{R}(\cdot)$ is the performance improvement induced by the design $\pi(s_t;\theta)$. The policy $\pi_\theta$ is trained to maximize expected performance over $\mathcal{P}$, directly addressing the expert dependence of hand-crafted optimizers.

Recent years have seen a broad spectrum of MetaBBO algorithms, which fall into four lines by training strategy: 1) \emph{RL-based}, modeling algorithm design as a Markov Decision Process~\citep{tianye-1,tianye-2,designx,gleet,rldas} and training the meta-level policy with Q-learning~\citep{mnih2013playing} or Policy Gradient~\citep{schulman2017proximal}; 2) \emph{pretrained optimization models}, representing optimization dynamics with neural networks trained by supervised~\citep{glhf,pom-2} or unsupervised~\citep{wangchao-1, wangchao-2,wangchao-3} performance signals; 3) \emph{LLM-based}, adopting LLMs as the meta-level policy and exploiting their in-context learning for iterative design~\citep{romeraparedes2024funsearch, novikov2025alphaevolve, liu2024eoh, ye2024reevo, llamea,survey-niki,llameabo}; and 4) \emph{neuroevolution-based}, replacing RL with an evolution optimizer to train the meta-level policy~\citep{lange-lga,lange-les,lange-lqd, metade, autopso}. Beyond single-objective problems, MetaBBO has been extended to expensive~\citep{duyukun-1,duyukun-2,cobra++}, multi-objective~\citep{metabbo-as-2}, dynamic~\citep{gaozijian}, multitask~\citep{l2t,metamto}, and large-scale optimization~\citep{qiuwenjie,qiuwenjie-2,learn-to-decompose}, alongside efforts on optimization state learning~\citep{heike,heike-2,deepela,neurela,tianye-1}, benchmarking platforms~\citep{metabox,metabox2,liu2025llm4ad}, and test case generation~\citep{wangchen-1,wangchen-2,skvorc2026llm}. \textbf{These algorithms, however, still more or less demand deep expertise in designing their meta-level policies or low-level optimizers, which motivates us to automate the design of MetaBBO itself.}

\subsection{Openendedness}
Open-endedness originates in artificial life, aiming to reproduce in silico the unbounded novelty generation of biological evolution~\citep{open-survey,open-jeff,open-tim}. The Gödel machine~\citep{godel-1,godel-2} first conceptualized a fully self-referential solver that rewrites its own code once an optimal proof searcher proves the rewrite increases expected utility, yet a feasible implementation remained elusive until the advent of powerful LLM agents. Evolution through Large Models (ELM) uses code-trained LLMs as intelligent mutation operators to bootstrap programs into new domains~\citep{open-stanley-2}, with similar ideas in OMNI and OMNI-EPIC for human-aligned interestingness~\citep{open-omni-1,open-omni-2} and Voyager in embodied intelligence~\citep{open-voya}. However, the agent in such systems does not recursively rewrite its own improvement machinery, tightening the potential exploratory space. Later works such as the Darwin Gödel Machine~\citep{open-DGM} relaxes the proof requirement to empirical validation. Recently, Hyperagents~\citep{open-hyperagents} and Gödel Agent~\citep{recursive-2} further make the meta-level modification procedure itself editable, yielding fully open-ended systems with cross-domain performance gains. For a more detailed understanding, we suggest the survey on recursive self-improvement systems~\citep{rsi-survey}. HADA is developed under the umbrella of such recursive systems.

\section{Methodology}\label{sec:3}
As we discussed before, to introduce openendedness into design process of a MetaBBO algorithm, we propose Hyper Algorithm Design Agent~(HADA), which is based on LLM coding agents and allows open-ended self-improvement. In this section, we detail each algorithmic component of HADA, and show how HADA runs its workflow in a simple and elegant way, with minimal human inputs. 

\subsection{Overall Workflow}\label{sec:3.1}

The overall workflow of HADA is illustrated in Fig.~\ref{fig:intro}.
For completeness, we further provide the pseudocode of HADA in Alg.~\ref{alg:hada}.
HADA takes two types of inputs:
1) a \textbf{target MetaBBO project}, which comprises the source code of key components in a MetaBBO algorithm, including the meta-level policy, low-level optimizer, optimization problem set, training logic, and evaluation protocol---denoted as $\mathbb{M}$;
and 2) \textbf{hyper \& task agents}, i.e., two LLM-based coding agents $\mathbb{A}_{hyper}$ and $\mathbb{A}_{task}$ that specialize in open-ended improvement and task solving, respectively.
The output of HADA is a carefully evolved MetaBBO project $\mathbb{M}^*$, which is obtained through iterative evolutionary modifications driven by $\mathbb{A}_{hyper}$ and $\mathbb{A}_{task}$, thereby attaining the best achievable optimization performance on the corresponding tasks.
Notably, all inputs and outputs of HADA are treated as editable source code files.

At the initialization phase, HADA first instantiates an algorithm evolution tree $\mathbf{AET}$ which uses a tree-based structure to save MetaBBO evolution history. An initial MetaBBO project $\mathbb{M}^{(0)}$ is first evaluated by executing its sourcecodes, training the meta-level policy and testing the resulted optimization performance on the target benchmark. A structured in-execution information $\mathbf{Info}$ is then constructed to include current codes, previous patches, execution logs, performance scores etc. Then $\mathbb{M}^{(0)}$ and its $\mathbf{Info}$ are recorded by $\mathbf{AET}$. After the initialization, HADA triggers an open-ended MetaBBO design loop by first sampling a parent node from $\mathbf{AET}$~(the sampling strategy is detailed in next section). Given the codes and $\mathbf{Info}$ of the sampled parent project, hyper agent and task agent~($\mathbb{A}_{hyper}$ and $\mathbb{A}_{task}$) follow a sequential operation order to evolve HADA system: 


\textsc{Step 1}: the hyper agent $\mathbb{A}_{hyper}$ is granted full access to all editable files, including both the MetaBBO project and the agent files.
Conditioned on comprehensive in-execution information\footnote{In our implementation, HADA not only allows the hyper agent to decide modifications based on the sampled parent project, but also grants it access to all information in $\mathbf{AET}$.}, $\mathbb{A}_{hyper}$ reasons about and modifies the agent files that specify how the hyper \& task agents operate\footnote{In our implementation, we also leave the freedom of modification to the hyper agent: beyond the agent files, it may modify the MetaBBO project when necessary, though this rarely happens in practice.}.
A key aspect of this modification cycle is that the hyper agent is permitted to access and modify itself.
While modifying the task agent already unlocks potential improvements to the MetaBBO design, this self-modification opens the door to open-endedness, endowing HADA with a remarkable capability for novelty search.

\textsc{step 2}: Modified by $\mathbb{A}_{hyper}$, the task agent $\mathbb{A}_{task}$ is allowed to access only the MetaBBO project, and uses the modified thinking pattern to modify that project for potential improvement. Once this Hyper-to-Task pipeline ends, the resulted new MetaBBO project undergoes the same evaluation as $\mathbb{M}^{(0)}$, and $\mathbf{AET}$ inserts this project and its in-execution information into the evolution history, as the child of the parent project. This pipeline loops for $\mathcal{H}$ steps for continual recursive self-improvement. We next further elaborate the technical detail within this workflow.  

\begin{algorithm}[t]
\caption{Hyper Algorithm Design Agent}
\label{alg:hada}
\small
\KwIn{Initial MetaBBO project $\mathbb{M}^{(0)}$, hyper \& task agent $\{\mathbb{A}_{hyper},\mathbb{A}_{task}\}$, budget $\mathcal{H}$.}
\KwOut{Optimal MetaBBO project $\mathbb{M}^*$.}

\textcolor{commentgray}{\textit{/*Initialization*/}} \\
Initialize algorithm evolution tree: $\mathbf{AET} = \emptyset$ \;
Evaluate the in-execution information:  $\mathbf{Info} = \mathbb{M}^{(0)}.\text{evaluate()}$  \;
$\mathbf{AET}$ records the project: $\mathbf{AET}.\text{insert}(\mathbb{M}^{(0)}, \mathbf{Info})$ \;
\textcolor{commentgray}{\textit{/*Open-ended MetaBBO design loop*/}}\\
\For{$h = 1$ \KwTo $\mathcal{H}$}{
    Sample a parent project from history: $\mathbb{M}^{(h)}, \mathbf{Info}(\mathbb{M}^{(h)}) = \mathbf{AET}.\text{sample}()$ \;
    \textcolor{commentgray}{\textit{/*Hyper-to-Task editing workflow*/}} \\
    Hyper agent improves task agent and itself: $\{\mathbb{A}_{hyper},\mathbb{A}_{task}\} = \mathbb{A}_{hyper}.\text{modify}(\{\mathbb{A}_{hyper},\mathbb{A}_{task}\}|\mathbf{Info}(\mathbb{M}^{(h)})$\;
    Task agent improves MetaBBO: $\mathbb{M}^{(h)} = \mathbb{A}_{task}.\text{modify}(\mathbb{M}^{(h)}|\mathbf{Info}(\mathbb{M}^{(h)}))$ \;
    \textcolor{commentgray}{\textit{/*Evaluate the modified MetaBBO*/}}\\
    Evaluate the in-execution information:  $\mathbf{Info} = \mathbb{M}^{(h)}.\text{evaluate()}$  \;
    \textcolor{commentgray}{\textit{/*Update evolution history*/}}\\
    $\mathbf{AET}.\text{insert}(\mathbb{M}^{(h)}, \mathbf{Info})$ \;
}
\Return $\mathbf{AET}.\text{optimal}()$ \;
\end{algorithm}

\subsection{Design Components}
\textbf{MetaBBO Project.} 
In HADA, a MetaBBO project $\mathbb{M}$ is basically a formal \emph{Python} project like any project in your PyCharm or VS Code. Despite the cumbersome dependency files and package management files, to core of a MetaBBO project includes four types of files. According to existing standard MetaBBO benchmark platforms~\citep{metabox,metabox2}, these files are: 1) Meta-level policy, where the neural network architecture, inference logic, rollout pipeline of the policy are detailed; 2) Low-level optimization environment, which is the composition of an evolutionary optimizer and an optimization problem instance. As we described in Sec.~\ref{sec:2.1}, the low-level optimization dynamic is controlled by the meta-level policy through learning; 3)Training logic, which clarify how the meta-level policy is trained given the feedback signals from the low-level optimization, and also indicate the training problem set~($\mathcal{P}$ in Eq.~(\ref{eq:metabbo})); 4) Testing procedure, which evaluate the optimization performance of the trained MetaBBO on the testing problem set. For each tested instance, normally multiple independent runs are needed to reduce experimental variance. HADA allows the coding agents possess holistic perception field and operational permission, which is particularly superior to human experts when the project is huge.

\textbf{In-execution Information.} In each evolution step $h$, once the two coding agents finishes the modification on the sampled parent MetaBBO project, a new child project $\mathbb{M}^{(h)}$ is created. To attain a comprehensive and objective feedback that could reflect how much the design of the MetaBBO project is improved, we stipulate a dictionary-like in-execution information object $\mathbf{Info}$, which thoroughly profiles the timely state of $\mathbb{M}^{(h)}$. Specifically, $\mathbf{Info}$ includes: 1) The current sourcecodes of $\mathbb{M}^{(h)}$; 2) All patches made by the coding agents; 3) Domain knowledge generated by the coding agents, which records the definition, problem property and solving experiences on the target optimization domain; 4) Evaluation logs that report intermediate logging data during the MetaBBO's training and testing; 5) Scores, which include a group of per-run scores $\{\{\mathbf{Perf}_{i,j}\}_{i=1}^{N}\}_{j=1}^{M}$ where the MetaBBO is tested across $N$ testing problems for $M$ independent runs, and an aggregated score $\overline{\mathbf{Perf}}$ averages these per-run scores. We leave the scoring detail in the next paragraph.

\textbf{Unified Performance Evaluation.} A key challenge for a universal optimization system is its compatibility across different optimization domains or problems. One can imagine that for two different problems, their optimal values, landscapes and objective scales are quite distinct. This issue may misleads the coding agents in HADA when they face different optimization tasks. To address this, we introduce an additional normalization trick. Specifically, suppose we are doing minimization, the per-run score $\mathbf{Perf}_{i,j}$ is computed as $\frac{f_{i,j}^{0} - f_{i,j}^{T}}{f_{i,j}^{0} - f_i^*}$, where $f_i^*$ is the optimal value of $i$-th testing problems, and $f_{i,j}^{t}$ is the best-so-far objective value at $t$-th optimization step. We scale the performance score to 0-1 for different target problems in different MetaBBO projects. For a MetaBBO project with syntax error or runtime error during the evaluation, we set its $\overline{\mathbf{Perf}}$ as $\text{NA}$.   

\textbf{Algorithm Evolution Tree.} The algorithm evolution tree $\mathbf{AET}$ resembles git management workflow with a simpler structure. When a newly modified MetaBBO project needs to be saved into $\mathbf{AET}$, HADA puts it under the parent project sampled before~(see a complete $\mathbf{AET}$ in Fig.~\ref{fig:AET}). To sample a parent project from $\mathbf{AET}$, we borrow the idea from Zhang et al.~\citeyearpar{open-DGM,open-hyperagents}, where parent selection is based on each agent's performance score and its number of children. This strategy focuses on the promising and less explored node, while addresses exploration \& exploitation tradeoff in general cases. Each node has a non-zero selection probability to ensure the search diversity. For those nodes with NA score, we do not allow sample them to avoid computational resource waste.

\textbf{Hyper \& Task Agent.} The hyper agent $\mathbb{A}_{hyper}$ and the task agent $\mathbb{A}_{task}$ are closely tied while serve for distinct roles. For the task agent, the core task is to follow the modification suggestions from the hyper agent and refine the sampled parent MetaBBO project correspondingly. For the hyper agent, its core task, instead, is to provide the openendedness for the whole HADA system by modifying not only the task agent~(how to improve) but also itself~(thinking of how to improve). We leave the prompts of $\mathbb{A}_{hyper}$ and $\mathbb{A}_{task}$ at Appendix~\ref{appx:prompt},where we show the initial prompts and final prompts after HADA's open-ended evolution. This recursive self-improvement in $\mathbb{A}_{hyper}$ and $\mathbb{A}_{task}$ also helps them reduce the risk of sensitive prompt~\citep{prompt-1,prompt-2}.

\section{Experimental Results}\label{sec:4}
\subsection{Experimental Settings}\label{sec:4.1}
\textbf{HADA.} In our main experiments, during the evolution process, HADA evaluates the MetaBBO project by training it with 5 epochs and testing the trained policy on test set for 5 independent runs~(serve as proxy evaluation for saving resources). After the evolution, the finally obtained MetaBBO project is trained for 20 epochs and tested for 10 independent runs~(serve as official evaluation). We set the evolution horizon of HADA as 100. We adopt DeepSeek‑V4‑Pro\footnote{\textcolor{orange}{\url{https://api.deepseek.com}}} as the LLM backbone for both the hyper agent and task agent, we set its maximal output length as 2e4.


\textbf{Testbeds.} The experiments involve four diverse optimization domains: 1) \emph{Single Objective Optimization}, where the 24 synthetic instances~(20D, 1e4 FEs) with random rotation and shift in COCO-BBOB testsuite~\citep{coco} are used as target problem distribution $\mathcal{P}$; 2) \emph{Constrained Optimization}, where the 54 synthetic nonlinear constrained instances~(20D, 1e4 FEs) in COCO-constrained testsuite~\citep{coco-constrained} are used. For these constrained problems, the per-run score $\mathbf{Perf}$ is set to 0 if no feasible solution is found, otherwise it is set to the objective value finally achieved; 3) \emph{Multi Objective Optimization}, where we use the nine 5-objective instances of WFG functions~\citep{wfg}~(WFG1-WFG9, 28D, 2e3 FEs) as the target problems, and use the normalized hypervolume as the per-score; 4) \emph{Realistic UAV Planning}, where we use a recently proposed benchmark~\citep{uav-bench} as the target problems. Specifically, we use its implementation in MetaBox-v2~\citep{metabox2} and instantiate 56 instances~(30D, 1.5e5 FEs). These highly constrained UAV path planning problems are transformed into single objective problem by weighted-sum trick in MetaBox-v2. The concrete train-test split for the mentioned testsuites can be found in our project.

\textbf{Baselines.} For single objective scenarios COCO-BBOB and UAV planning, we consider following baselines: human-crafted BBO algorithms DE~\citep{de}, SHADE~\citep{shade}, JDE21~\citep{jde21}, MADDE~\citep{madde} and CMAES~\citep{cmaes}; MetaBBO algorithms LDE~\citep{lde}, RL-DAS~\citep{rldas}, GLEET~\citep{gleet}. For constrained optimization, we compare human-crafted baselines L-SHADE-BOC~\citep{kawachi2019shade}, AL1-CMA-ES~\citep{Dufoss2022BenchmarkingSS}, BP-$\epsilon$MAg-ES~\citep{BP-epsMAg-ES}, MDE-CGO~\citep{MDE-CGO}; MetaBBO algorithm MeCO~\citep{MeCO} and LAMDE~\citep{LAMDE}. For multi-objective optimization, we compare human-crafted baselines GDE3~\citep{GDE3}, NSGAIII~\citep{NSGA3}, RVEA~\citep{RVEA}, SPEA2~\citep{SPEA2}, MOEA/D~\citep{MOEA/D}, R-MODE~\citep{R-MODE}; MetaBBO algorithms MADAC~\citep{madac}. We connect HADA with MetaBox-v2 to attain the implementation of these baselines. We also conducted hybrid search on their hyper-parameters to attain optimal performance for comparison, see Appendix~\ref{appx:parameter} for details. All experiments are performed on a machine with 8-core Intel(R) Xeon(R) Platinum CPU and 16GB RAM. 

\subsection{Algorithm Design Capability~(RQ1)}\label{sec:4.2}

\begin{table}[t]
    \centering
    \caption{Final performance on held-out BBOB functions (1/2).
    Each cell reports the mean $\pm$ standard deviation over independent runs. Column bests are shaded and bold;
    runners-up are underlined.}
    \label{tab:all_baselines_bbob}
    \scriptsize
    \setlength{\tabcolsep}{2.5pt}
    \renewcommand{\arraystretch}{1.25}

    \resizebox{0.9\textwidth}{!}{
    \begin{tabular}{@{}c@{\hspace{12pt}}l*{8}{c}@{}}
        \toprule
        & \textbf{Method}
        & $f_{3}$ & $f_{4}$ & $f_{6}$ & $f_{7}$
        & $f_{9}$ & $f_{13}$ & $f_{14}$ & $f_{16}$ \\
        \midrule
        \multirow{5}{*}{\rotatebox[origin=c]{90}{\textbf{BBO}}}
        & DE
        & \bbobinlineval{0.8773}{0.0355} & \bbobinlineval{0.9068}{0.0181} & \bbobinlinesecond{0.9999}{0.0000} & \bbobinlineval{0.9706}{0.0073}
        & \bbobinlineval{0.9997}{0.0001} & \bbobinlineval{0.9636}{0.0101} & \bbobinlineval{0.9981}{0.0006} & \bbobinlineval{0.4208}{0.0968} \\
        & SHADE
        & \bbobinlineval{0.8639}{0.0403} & \bbobinlineval{0.8898}{0.0186} & \bbobinlinesecond{0.9999}{0.0000} & \bbobinlineval{0.9884}{0.0047}
        & \bbobinlinesecond{0.9998}{0.0001} & \bbobinlineval{0.9810}{0.0073} & \bbobinlineval{0.9990}{0.0003} & \bbobinlineval{0.4201}{0.1345} \\
        & JDE21
        & \bbobinlinesecond{0.9705}{0.0110} & \bbobinlinesecond{0.9756}{0.0084} & \bbobinlinesecond{0.9999}{0.0000} & \bbobinlineval{0.9844}{0.0051}
        & \bbobinlineval{0.9997}{0.0001} & \bbobinlineval{0.9887}{0.0045} & \bbobinlinesecond{0.9992}{0.0006} & \bbobinlineval{0.4619}{0.1208} \\
        & MADDE
        & \bbobinlineval{0.8892}{0.0261} & \bbobinlineval{0.9151}{0.0208} & \bbobinlinebest{1.0000}{0.0000} & \bbobinlineval{0.9872}{0.0045}
        & \bbobinlineval{0.9997}{0.0001} & \bbobinlineval{0.9814}{0.0049} & \bbobinlineval{0.9967}{0.0014} & \bbobinlineval{0.6200}{0.1116} \\
        & CMAES
        & \bbobinlineval{0.9652}{0.0254} & \bbobinlineval{0.9665}{0.0074} & \bbobinlinebest{1.0000}{0.0000} & \bbobinlinebest{0.9993}{0.0011}
        & \bbobinlineval{0.8289}{0.1400} & \bbobinlinebest{0.9999}{0.0001} & \bbobinlinebest{1.0000}{0.0000} & \bbobinlineval{0.5169}{0.1912} \\
        \midrule
        \multirow{4}{*}{\rotatebox[origin=c]{90}{\textbf{MetaBBO}}}
        & RL-DAS
        & \bbobinlineval{0.9362}{0.0104} & \bbobinlineval{0.9098}{0.0167} & \bbobinlinebest{1.0000}{0.0000} & \bbobinlineval{0.9971}{0.0006}
        & \bbobinlinebest{0.9999}{0.0000} & \bbobinlineval{0.9452}{0.0040} & \bbobinlineval{0.9970}{0.0013} & \bbobinlineval{0.6296}{0.0596} \\
        & DQN-DE
        & \bbobinlineval{0.9348}{0.0462} & \bbobinlineval{0.9751}{0.0192} & \bbobinlinesecond{0.9999}{0.0000} & \bbobinlineval{0.9557}{0.0199}
        & \bbobinlineval{0.9981}{0.0019} & \bbobinlineval{0.9478}{0.0091} & \bbobinlineval{0.9724}{0.0229} & \bbobinlineval{0.6620}{0.0446} \\
        & LDE
        & \bbobinlineval{0.8668}{0.0389} & \bbobinlineval{0.8644}{0.0279} & \bbobinlinesecond{0.9999}{0.0000} & \bbobinlineval{0.9869}{0.0061}
        & \bbobinlineval{0.9994}{0.0002} & \bbobinlineval{0.9249}{0.0143} & \bbobinlineval{0.9899}{0.0046} & \bbobinlineval{0.6005}{0.0648} \\
        & GLEET
        & \bbobinlineval{0.8578}{0.0429} & \bbobinlineval{0.8707}{0.0137} & \bbobinlinesecond{0.9999}{0.0001} & \bbobinlineval{0.9837}{0.0082}
        & \bbobinlineval{0.9995}{0.0003} & \bbobinlineval{0.9794}{0.0167} & \bbobinlineval{0.9967}{0.0045} & \bbobinlinesecond{0.8409}{0.0474} \\
        \midrule
        \multirow{1}{*}{\rotatebox[origin=c]{90}{\textbf{ }}}
        & HADA
        & \bbobinlinebest{0.9938}{0.0032} & \bbobinlinebest{0.9925}{0.0027} & \bbobinlinebest{1.0000}{0.0000} & \bbobinlinesecond{0.9979}{0.0015}
        & \bbobinlinebest{0.9999}{0.0000} & \bbobinlinesecond{0.9985}{0.0016} & \bbobinlinebest{1.0000}{0.0000} & \bbobinlinebest{0.8631}{0.0696} \\
        \bottomrule
    \end{tabular}
    }
\end{table}

\begin{table}[t]
    \centering
    \caption{Final performance on held-out set (2/2), last column denotes average across all.}
    \label{tab:all_baselines_bbob_part2}
    \scriptsize
    \setlength{\tabcolsep}{2.5pt}
    \renewcommand{\arraystretch}{1.25}
    \resizebox{\textwidth}{!}{
    \begin{tabular}{@{}c@{\hspace{12pt}}l*{9}{c}@{}}
        \toprule
        & \textbf{Method}
        & $f_{17}$ & $f_{18}$ & $f_{19}$ & $f_{20}$
        & $f_{21}$ & $f_{22}$ & $f_{23}$ & $f_{24}$ & Avg.\\
        \midrule
        \multirow{5}{*}{\rotatebox[origin=c]{90}{\textbf{BBO}}}
        & DE
        & \bbobinlineval{0.9338}{0.0178} & \bbobinlineval{0.9095}{0.0143} & \bbobinlineval{0.7691}{0.0256} & \bbobinlinesecond{0.9999}{0.0000}
        & \bbobinlineval{0.9041}{0.0575} & \bbobinlineval{0.9386}{0.0742} & \bbobinlineval{0.4811}{0.1418} & \bbobinlineval{0.7134}{0.0275} & \bbobinlineval{0.8618}{0.0141} \\
        & SHADE
        & \bbobinlineval{0.8935}{0.0309} & \bbobinlineval{0.8983}{0.0207} & \bbobinlineval{0.7686}{0.0330} & \bbobinlinesecond{0.9999}{0.0000}
        & \bbobinlineval{0.9322}{0.0579} & \bbobinlinebest{0.9755}{0.0004} & \bbobinlineval{0.4063}{0.1562} & \bbobinlineval{0.7233}{0.0328} & \bbobinlineval{0.8594}{0.0132} \\
        & JDE21
        & \bbobinlineval{0.8827}{0.0411} & \bbobinlineval{0.9157}{0.0405} & \bbobinlineval{0.7351}{0.0368} & \bbobinlinebest{1.0000}{0.0000}
        & \bbobinlineval{0.9424}{0.0596} & \bbobinlineval{0.9726}{0.0063} & \bbobinlineval{0.4567}{0.0799} & \bbobinlineval{0.6997}{0.0315} & \bbobinlineval{0.8741}{0.0129} \\
        & MADDE
        & \bbobinlineval{0.8541}{0.0407} & \bbobinlineval{0.8757}{0.0253} & \bbobinlineval{0.7887}{0.0294} & \bbobinlinesecond{0.9999}{0.0000}
        & \bbobinlinebest{0.9963}{0.0060} & \bbobinlinesecond{0.9746}{0.0009} & \bbobinlineval{0.4601}{0.1062} & \bbobinlineval{0.6555}{0.0243} & \bbobinlineval{0.8748}{0.0138} \\
        & CMAES
        & \bbobinlinebest{0.9994}{0.0005} & \bbobinlinebest{0.9990}{0.0006} & \bbobinlineval{0.0415}{0.0364} & \bbobinlinesecond{0.9999}{0.0000}
        & \bbobinlineval{0.9395}{0.0750} & \bbobinlineval{0.9602}{0.0454} & \bbobinlineval{0.5451}{0.1624} & \bbobinlineval{0.6221}{0.0314} & \bbobinlineval{0.8342}{0.0113} \\
        \midrule
        \multirow{4}{*}{\rotatebox[origin=c]{90}{\textbf{MetaBBO}}}
        & RL-DAS
        & \bbobinlineval{0.8955}{0.0074} & \bbobinlineval{0.8889}{0.0150} & \bbobinlineval{0.8188}{0.0148} & \bbobinlinebest{1.0000}{0.0000}
        & \bbobinlineval{0.8635}{0.0438} & \bbobinlineval{0.9570}{0.026} & \bbobinlineval{0.4872}{0.2016} & \bbobinlineval{0.7172}{0.0229} & \bbobinlineval{0.8779}{0.0131} \\
        & DQN-DE
        & \bbobinlineval{0.9565}{0.0321} & \bbobinlineval{0.7772}{0.0444} & \bbobinlineval{0.7965}{0.0278} & \bbobinlinebest{1.0000}{0.000}
        & \bbobinlinesecond{0.9454}{0.0195} & \bbobinlineval{0.9584}{0.0492} & \bbobinlineval{0.5748}{0.1505} & \bbobinlineval{0.6730}{0.042} & \bbobinlineval{0.8831}{0.0124} \\
        & LDE
        & \bbobinlineval{0.8588}{0.0359} & \bbobinlineval{0.8676}{0.0271} & \bbobinlineval{0.7795}{0.0411} & \bbobinlinesecond{0.9999}{0.0000}
        & \bbobinlineval{0.9218}{0.0512} & \bbobinlineval{0.9719}{0.0045} & \bbobinlineval{0.5062}{0.1027} & \bbobinlineval{0.6945}{0.0309} & \bbobinlineval{0.8652}{0.0053} \\
        & GLEET
        & \bbobinlineval{0.8491}{0.0469} & \bbobinlineval{0.8316}{0.0562} & \bbobinlinesecond{0.8200}{0.0337} & \bbobinlinesecond{0.9999}{0.0000}
        & \bbobinlineval{0.9346}{0.0680} & \bbobinlineval{0.9432}{0.0662} & \bbobinlinesecond{0.5924}{0.1216} & \bbobinlinesecond{0.8017}{0.0508} & \bbobinlinesecond{0.8942}{0.0086} \\
        \midrule
        \multirow{1}{*}{\rotatebox[origin=c]{90}{\textbf{ }}}
        & HADA
        & \bbobinlinesecond{0.9953}{0.0023} & \bbobinlinesecond{0.9818}{0.0096} & \bbobinlinebest{0.9382}{0.0326} & \bbobinlinebest{1.0000}{0.0000}
        & \bbobinlineval{0.9358}{0.0823} & \bbobinlineval{0.9420}{0.1079} & \bbobinlinebest{0.9189}{0.0257} & \bbobinlinebest{0.9175}{0.0229} & \bbobinlinebest{0.9674}{0.0078} \\
        \bottomrule
    \end{tabular}
    }
\end{table}

We validate whether HADA truly enables open-ended algorithm design in this section. Specifically, we focus on single-objective optimization scenario and prepare a naive MetaBBO backbone as the initial MetaBBO project $\mathbb{M}^{(0)}$ for HADA. Its meta-level is a DQN~\cite{mnih2013playing} policy that simply configures $F$ and $Cr$ of low-level DE optimizer. We term this backbone as DQN-DE and provide its full details at Appendix~\ref{appx:dqnde}. Table~\ref{tab:all_baselines_bbob} and Table~\ref{tab:all_baselines_bbob_part2} present the final per-run performance scores of the best MetaBBO project obtained from HADA and the baseline algorithms across the testing instances in COCO-BBOB set. Following key observations can be concluded: 

1) Overall (see the last column in Table~\ref{tab:all_baselines_bbob_part2}), MetaBBOs generally outperform handcrafted BBOs, confirming that meta-learning mitigates the expertise needs of traditional BBOs. More importantly, HADA achieves a significant performance leap over existing MetaBBOs: using the worst-performing CMAES as baseline, HADA improves upon the SOTA MetaBBO (GLEET) by 121.7\%, which we attribute to its efficient automated workflow and open-ended algorithm evolution. Furthermore, HADA consistently outperforms LLaMEA~\citep{llamea}, the SOTA LLM-based algorithm design framework (results in Appendix~\ref{appx:ablation}, Fig.~\ref{fig:ablation} due to space limit).; 

2) We can also observe that either the BBOs and the MetaBBOs show biased performance on different problems. Such performance distribution imbalance exactly reflect the subjectivity of their human-based designs behind. The developers of these algorithms easily introduce design bias based on their own experiences. Instead, HADA's automated self-improvement loop ensures an objective and comprehensive search. Hence, HADA achieves more generally good performance; 

3) We especially would like to discuss the optimization under challenging cases. It can be observed both BBOs and MetaBBOs perform relatively bad on $f_{19}$, $f_{23}$ and $f_{24}$, which are Griewank-Rosenbrock, Katsuura and Lunacek bi-Rastrigin problems. These problems feature highly compositional landscapes that challenge the learning capability of MetaBBOs. In such situation, HADA is still capable of evolving a MetaBBO variant with robust optimization performance, this is a clear evidence for the open-ended potential in HADA.

\subsection{Generalization Test~(RQ2)}
\begin{figure}[t]
    \centering   
    \includegraphics[width=0.8\textwidth]{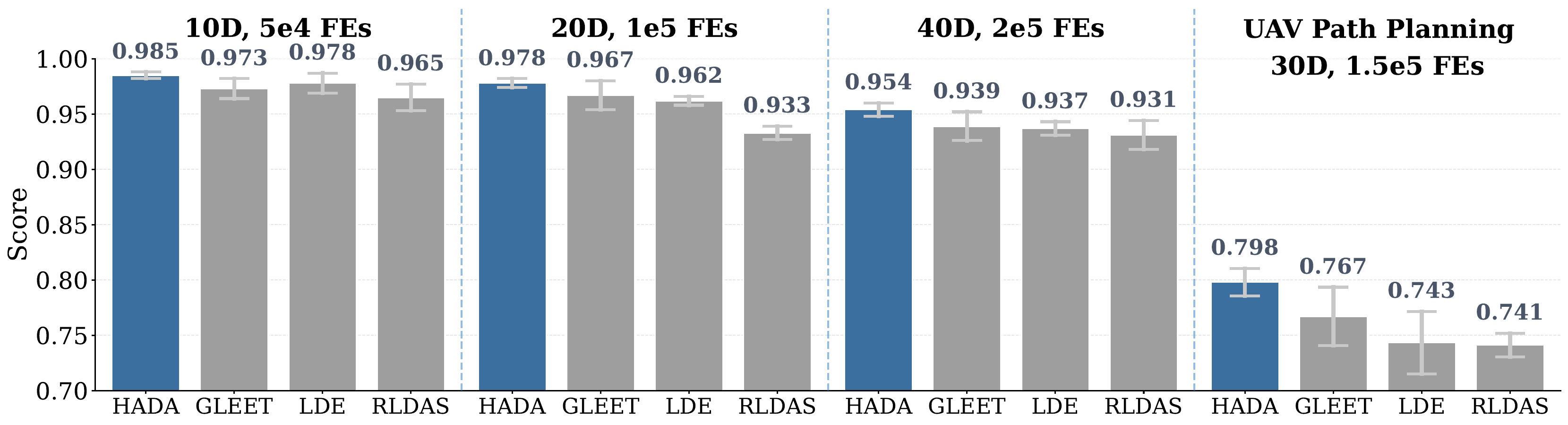}
    \caption{Out-of-distribution generalization comparison under diverse scenarios.}
    \label{fig:generalization}
\end{figure}
Recall that the core motivation of MetaBBO researches is to enhance the generalization ability across different problems. Due to this, it is necessary to compare the generalization performance of HADA and existing MetaBBO baselines, to validate the MetaBBO variant proposed by HADA does not sacrifice general solving ability for overfitting a specific problem settings. To this end, we test our HADA and the other three MetaBBO baselines~(trained in Sec.~\ref{sec:4.2}) on four different problem settings: three COCO-BBOB settings and a realistic UAV path planning scenario. In this case, the meta-level policies in the baselines are directly zero-shot to the test set without fine-tuning. We report in Fig.~\ref{fig:generalization} the averaged performance score $\overline{\mathbf{Perf}}$ on the four different out-of-distribution generalization settings. The results demonstrate that HADA's open-ended design does not overfit easily.

\subsection{Interpretation Analysis~(RQ3)}

Another key research question is how to open the ``black-box'' of HADA. That is, given the state-of-the-art performance achieved by HADA in the previous two sections, what is the core thinking and steps HADA uses to design MetaBBO algorithm? In this section, we look into this by reviewing the evolution process of HADA on single-objective optimization scenario. We illustrate the complete algorithm evolution tree $\mathbf{AET}$ during the open-ended process in Fig.~\ref{fig:AET}, where \#xx denotes the evolution step a node is saved into $\mathbf{AET}$ and the numerical value is the corresponding $\overline{\mathbf{Perf}}$. We abstract several key nodes in this tree to interpret HADA's design philosophy:

\begin{figure}[h]
    \centering   
    \includegraphics[width=0.9\textwidth]{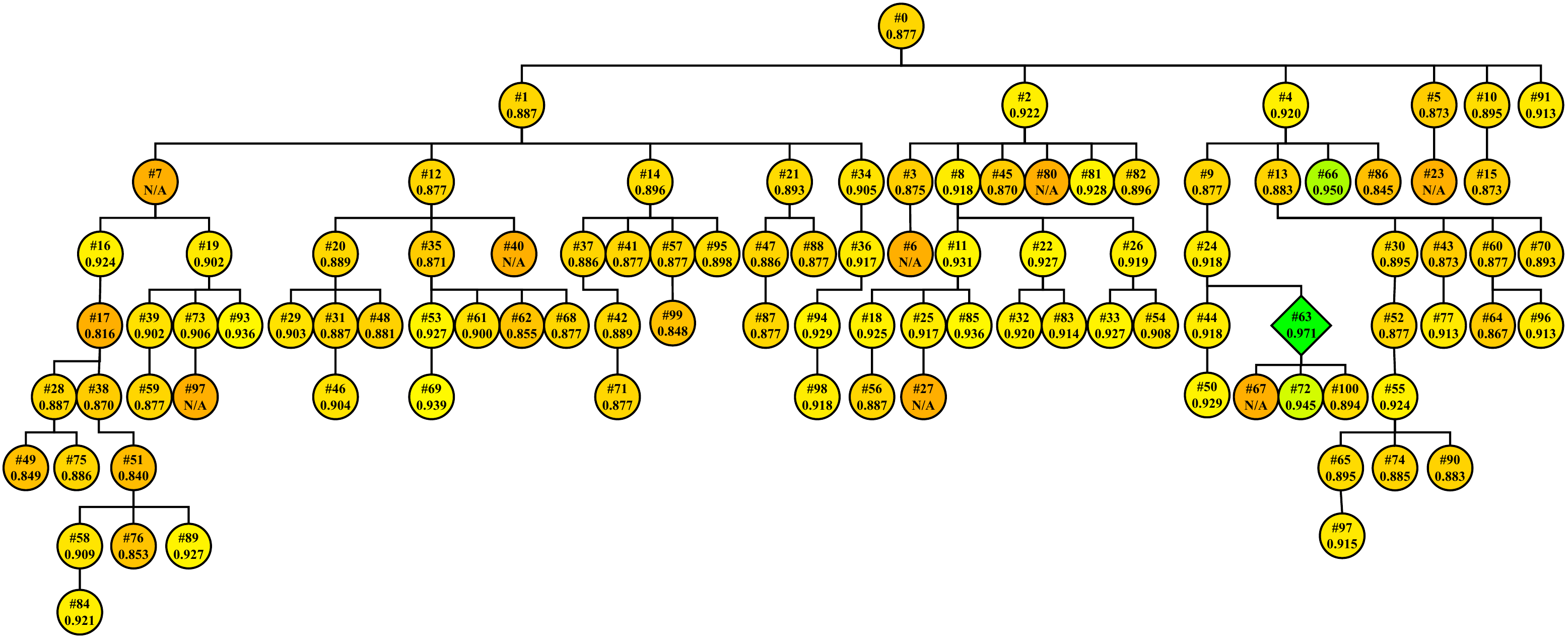}
    \caption{A complete algorithm evolution tree for single-objective domain.}
    \label{fig:AET}
\end{figure}

1) Step~\#4 ($0.877 \to 0.920$): The hyper agent first revises the task agent's
prompt to enforce low-level optimizer replacement (with \emph{domain guidance}),
encourage more substantial structural adjustments, and specify four
implementation templates together with a standard verification protocol. The
task agent then modifies the optimizer accordingly, yielding a SHADE-like
variant.

2) Steps~\#9, \#24 ($0.920 \to 0.877 \to 0.918$): At these steps, the
performance of the searched MetaBBO variants falls short of expectations.
In response, the hyper agent strictly prohibits the task agent from making
minor or irrelevant code modifications, and attempts to formally define what
constitutes structural novelty in an optimizer. It further refines its own
instructions by adding explicit principles for proposing novel algorithms.
The task agent then acts accordingly: it first rolls back (\#9) and
subsequently explores a new direction (\#24).

3) Steps~\#19, \#49, \#53: In these intermediate steps, HADA shifts its focus
toward meta-level learning design within the MetaBBO project, e.g.,
normalization tricks for optimization-state scale stability, optimization-state
augmentation to support diverse optimization behaviors, and modifications to
the DQN policy network to enlarge the algorithm design space.

4) Step~\#63 ($0.918 \to 0.971$): Building on the explorations of all previous
steps, HADA reaches an ``Aha Moment'' at this step. The hyper agent first
modifies itself to enforce systematic inspection of patch files and the
detection of genuinely novel designs. It then rewrites the task agent's prompt
with two new directions: the meta-level policy should incorporate optimization
progress information to reduce learning difficulty, and the low-level
optimizer should integrate local search with different optimizers, such as PSO
or ES variants.

From this detailed analysis, we observe that HADA benefits from its
open-ended code-editing ability and progressively improves both the MetaBBO
algorithm and the agent's own reasoning through deliberate decisions. We
release the complete evolution logs in our source code and welcome further
analysis of this intriguing data.

\subsection{Cross-Domain Adaption~(RQ4)}
\begin{figure}[h]
    \centering   
    \includegraphics[width=0.8\textwidth]{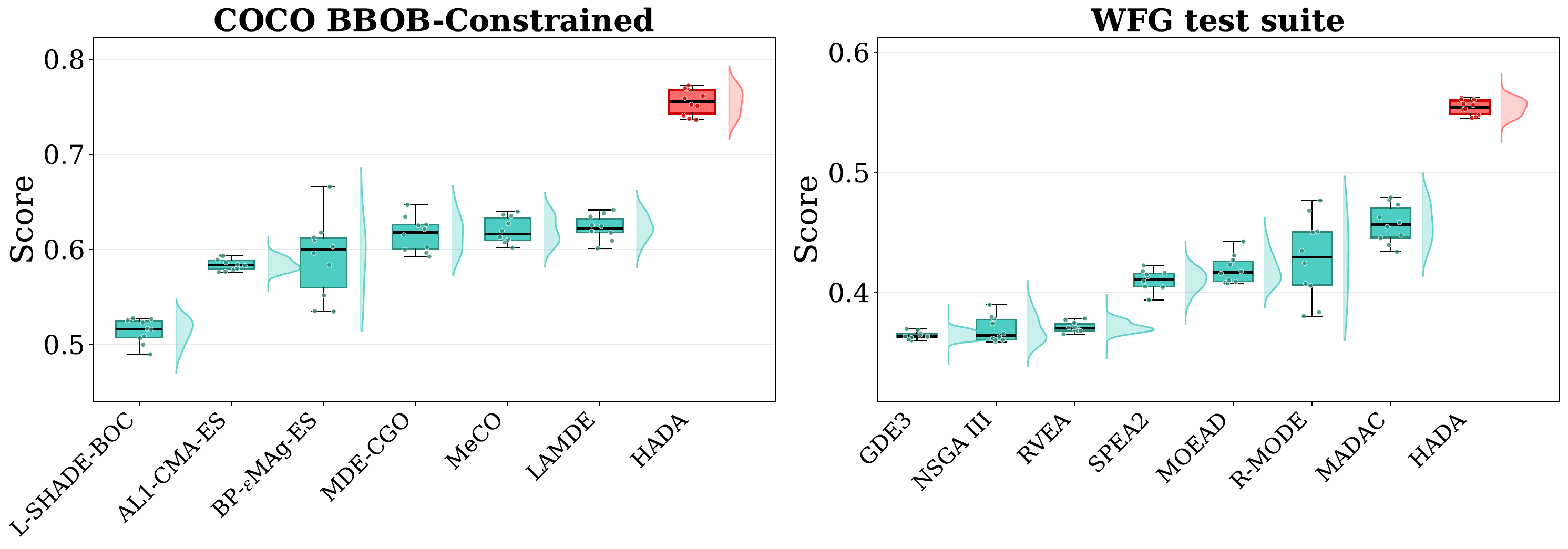}
    \caption{Cross-domain adaption results of HADA on two diverse scenarios.}
    \label{fig:migration}
\end{figure}

In this section, we evaluate HADA's adaptability beyond the single-objective domain studied so far. We select two representative optimization domains: constrained optimization~\citep{coco-constrained} and multi-objective optimization~\citep{wfg}, and run the HADA evolution loop~(Alg.~\ref{alg:hada}) under the settings of Sec.~\ref{sec:4.1}, initializing the MetaBBO project $\mathbb{M}^{(0)}$ as the optimal $\mathbb{M}^{*}$ from Sec.~\ref{sec:4.1} to reflect knowledge transfer. Fig.~\ref{fig:migration} reports the mean and std of per-run scores $\mathbf{Perf}$ for HADA and the BBO/MetaBBO baselines. The results show that HADA's knowledge on designing single-objective algorithms transfers positively to these domains, and with proper adaptation, the resulting MetaBBO variant can even outperform state-of-the-art domain-specific baselines.

Due to the space limitation, we provide several ablation studies on our HADA to demonstrate our design choices are proper, which can be found at Appendix~\ref{appx:ablation}.



\section{Conclusion}
To summarize this paper, we would like to first clarify that this paper is well motivated by two key aspects: 1) With various researches applying LLMs for optimization problem solving, LLMs' open-ended potential in such tasks is under-explored; 2) More importantly, given the generalization potential of learning-assisted optimization techniques such as MetaBBO, its dependence on human-based design remains a problem. To this end, we propose HADA as an initial exploration to introduce openendedness into MetaBBO's design process. With the advanced coding capability in recent LLM agents, HADA adopts a bi-agent system to achieve recursive design improvement, where a task agent aims to improve MetaBBO's design, a hyper agent is allowed to modify the thinking logic of the task agent and itself to recursively improve the thinking of how to improve. We also carefully engineer the evolution history management and unified evaluation interface to ensure effective algorithm discovery and versatility in practice respectively. Through systematic experiments, we demonstrate that openendedness is truly a key to novel MetaBBO design. Nevertheless, HADA shows several promising future improvements. First, at its current version, we set HADA's searching behavior by the simple heuristic rule that balances the tradeoff between exploitation and exploration.  Future work could explore more recent alternatives~\citep{alphago,ding2025dynamic}, or let the hyper agent creates new ones. Second, in this paper we mainly focus on the usage of HADA in continuous optimization domains, a very interesting future work is to explore HADA on learning-assisted combinatorial optimization techniques such as Neural Combinatorial Optimization~(NCO)~\citep{nco-3,nco-1,nco-2}. In the end, we authors would like to sincerely appreciate the emergence of the current agent world, which makes us long for a wonderful future of optimization.

\bibliography{iclr2027_conference}
\bibliographystyle{iclr2027_conference}

\newpage
\appendix
\section{Detailed Formulation of the Baseline DQN-DE Algorithm}\label{appx:dqnde}
Our proposed HADA evolution framework is built upon the DQN-controlled Differential Evolution (DQN-DE) algorithm. This baseline algorithm integrates a classic differential evolution (DE) optimizer with a deep Q-network (DQN), which dynamically adjusts the core hyperparameters of DE during the optimization process. We provide the complete and detailed formulation of the DQN-DE algorithm and the unified performance evaluation metric in this appendix for reproducibility.

\subsection{Basic Optimization Framework}
We adopt the classic \textbf{DE/rand/1/bin} strategy as the basic optimization paradigm. The population size is set to $N=5d$, where $d$ denotes the dimension of the optimization problem. At each generation $t$, the DQN controller first perceives the current optimization state and outputs a set of adaptive DE parameters, including the mutation factor $F_t$ and crossover rate $CR_t$. The DE optimizer then utilizes these dynamic parameters to generate trial vectors and update the population.
\subsection{State Representation}
The state vector $s_t \in \mathbb{R}^3$ is constructed from the optimization trajectory:
\[
s_t = \left[ p_t,\ \rho_t,\ \tilde{f}^{t}_{i,j} \right]
\]
where $i$ denotes the $i$-th test optimization problem, $j$ denotes the $j$-th independent run, and each component is defined as follows.

\textbf{Normalized optimization progress}:
\[
p_t = \min\left(1, E_t / E_{\text{max}}\right)
\]
where $E_t$ is the number of consumed function evaluations at step $t$, and $E_{\text{max}}$ is the maximum evaluation budget for each optimization task.

\textbf{Relative fitness improvement}:
\[
\rho_t = \max\left(0, \min\left(1, \frac{\log(1+|f^{0}_{i,j}|) - \log(1+|f^{t}_{i,j}|)}{\max\left(1, \log(1+|f^{0}_{i,j}|)\right)}\right)\right)
\]
where $f^{0}_{i,j}$ is the initial objective value, and $f^{t}_{i,j}$ is the best‑so‑far objective value at the $t$-th optimization step for the $j$-th run on the $i$-th test problem.

\textbf{Normalized best fitness}:
\[
\tilde{f}^{t}_{i,j} = \tanh\left( \frac{\log_{10}(1+|f^{t}_{i,j}|) \cdot \mathrm{sign}(f^{t}_{i,j})}{10} \right)
\]
This compresses unbounded fitness into a bounded range and avoids undefined logarithm when $f^{t}_{i,j}=0$.
\subsection{Q-Network Architecture}
We employ a three-layer fully connected feedforward neural network as the Q-network. Let $h_0 = s_t$ denote the input layer. The hidden layer computation for $l=1,2,3$ is formulated as
\[
h_l = \LeakyReLU(W_l h_{l-1} + b_l), \quad \alpha=0.2
\]
The final Q-value for state-action pair $(s_t, a)$ is output by the linear projection layer:
\[
Q(s_t, a) = W_{\text{out}} h_3 + b_{\text{out}}
\]
The detailed layer dimensions and parameter statistics are summarized in Table \ref{tab:dqn_arch}. All network weights are initialized via the Kaiming uniform initialization, and all biases are initialized to zero. The training network and target network share the same initialization seed to ensure full reproducibility.

\begin{table}[h]
\centering
\caption{Network architecture and parameter statistics of the DQN controller.}
\label{tab:dqn_arch}
\begin{tabular}{cccc}
\toprule
Layer & Input Dim & Output Dim & Parameters \\
\midrule
fc1   & 3    & 128  & $3 \times 128 + 128 = 512$ \\
fc2   & 128  & 128  & $128 \times 128 + 128 = 16512$ \\
fc3   & 128  & 128  & $128 \times 128 + 128 = 16512$ \\
out   & 128  & 25   & $128 \times 25 + 25 = 3225$ \\
\midrule
\textbf{Total} & -- & -- & \textbf{36761} \\
\bottomrule
\end{tabular}
\end{table}
\subsection{Discrete Action Space}
To adapt the DQN discrete decision paradigm, we discretize the continuous DE hyperparameters into a finite action space. The mutation factor $F \in [0.1, 1.0]$ and crossover rate $CR \in [0.0, 1.0]$ are uniformly divided into $K=5$ intervals respectively, yielding $|A|=K^2=25$ discrete candidate actions. The discrete action sets are
\[
F \in \{0.1, 0.325, 0.55, 0.775, 1.0\}, \quad CR \in \{0.0, 0.25, 0.5, 0.75, 1.0\}
\]
\subsection{DQN Training Details}
We adopt online training with experience replay to optimize the DQN controller. During optimization, each transition tuple $(s_t, a_t, R_t, s_{t+1}, \text{done})$ is stored in a replay buffer with a maximum capacity of 1000. We design a binary reward function to reflect the optimization improvement:
\[
R_t =
\begin{cases}
1, & f^{t}_{i,j} < f^{t-1}_{i,j} \\
0, & \text{otherwise}
\end{cases}
\]
The network is updated every 10 steps by sampling a mini-batch of 64 transitions from the replay buffer. We minimize the temporal difference (TD) loss function:
\[
\mathcal{L}(\theta) = \mathbb{E}_{(s,a,R,s') \sim \mathcal{D}} \left[ \left( Q_\theta(s,a) - y \right)^2 \right]
\]
where the target value is defined as
\[
y = R + \gamma (1-\text{done}) \max_{a'} Q_{\theta^-}(s', a').
\]
The discount factor $\gamma$ is set to 0.99. The target network parameters $\theta^-$ are synchronized with the online network parameters $\theta$ every 50 training steps.

Additional training hyperparameters are set as follows: gradient clipping with a maximum norm of 1.0, Adam optimizer with a fixed learning rate of $10^{-4}$. For action selection, we adopt the $\epsilon$-greedy strategy, where $\epsilon$ decays linearly from $1.0$ to $0.05$ with a decay rate of $0.999$ per update. During inference, $\epsilon$ is set to $0$ for pure greedy decision-making.

The controller is trained for 5 epochs. In each epoch, the model interacts with all training tasks sequentially. Model weights and training statistics (loss, reward) are recorded after each epoch, and the model from the final epoch is used for testing.

\section{Agent Prompt}\label{appx:prompt}
\subsection{Hyper Agent Prompt}
The complete prompt given to the Hyper Agent at the first generation of HADA is listed below.
The Hyper Agent is responsible for improving the Task Agent's prompt and code-logic across generations.

\begin{promptbox}{Hyper Agent Prompt}
\begin{lstlisting}
You are a Hyper Agent that improves the task agent's performance.{domain_info}

## Codebase Structure (/hada/metabbo/):

The task agent modifies code in two main layers:

### 1. Evolutionary Algorithm Layer (ec_algorithm.py)
- ec_algorithm.py: The main evolutionary algorithm implementation (PSO, DE, CMA-ES, etc.). Contains the optimizer class that handles population initialization, iteration loop, solution evaluation, and result tracking.
- param_controller.py: Parameter controller interface. Defines the abstract interface for dynamic parameter adjustment.

### 2. Meta-Learning Layer (meta_learning.py, meta_learning_env.py, train_meta_learning.py)
- meta_learning.py: Meta-learning controller that dynamically adjusts algorithm parameters during optimization.
- meta_learning_env.py: Environment wrapper for training the meta-learning controller.
- train_meta_learning.py: Training scripts for the meta-learning controller.

ALLOWED FILES TO MODIFY:
- task_agent.py (prompt, logic)
- hyper_agent.py (this file - you can modify your own prompt/logic)
- New helper modules (task agent can import them)

PROTECTED FILES (DO NOT MODIFY):
- config.py, domains/, agent/, utils/, generate_loop.py, harness.py, report.py

PATH STRUCTURE:
- Code: /hada/
- Previous gen results: {eval_path}/gen_N/ (e.g., gen_1/ for first generation)
- Your output: /hada/agent_output/

WHAT TO ANALYZE (focus on the latest generation):
- {eval_path}/gen_N/generate.log - Look for errors, timeouts, "[TRAIN] Run X failed:"
- {eval_path}/gen_N/<domain>_eval/agent_evals/all_patch.diff - See what Task Agent changed
- {eval_path}/gen_N/<domain>_eval/agent_evals/chat_history_task_agent.md - See Task Agent's reasoning
- {eval_path}/gen_N/report.json - Evaluation scores

CRITICAL REQUIREMENTS - YOU MUST FOLLOW THESE EXACTLY:

1. YOU MUST MODIFY CODE FILES - Use the editor tool with command='str_replace' to modify files
2. DO NOT JUST VIEW FILES - You must make ACTUAL CODE CHANGES using str_replace
3. MANDATORY MODIFICATIONS - You MUST modify at least one of these files:
   - /hada/task_agent.py (improve the prompt so Task Agent actually modifies EC code)
   - /hada/hyper_agent.py (improve your own prompt/logic)

4. REQUIRED STEP-BY-STEP PROCESS:
   Step 1: Use editor view command to see the current code/logs
   Step 2: Use editor str_replace command to make improvements (YOU MUST DO THIS)
   Step 3: Verify your changes were applied
   Step 4: Respond with JSON

## IMPORTANT GUIDANCE FOR TASK AGENT:

[CRITICAL: The Task Agent MUST try different evolutionary algorithms. This is the #1 priority.]

Analysis of previous experiments shows that Task Agents consistently ONLY make small PSO parameter tweaks (adjusting w, c1, c2, adding turbulence, changing initialization) and NEVER replace the algorithm entirely. This has caused scores to plateau for 20+ generations.

You MUST ensure the task_agent.py prompt encourages the Task Agent to:

1. Try different evolutionary algorithms - not just parameter tweaks:
   - For bbob_unconstrained: DE, CMA-ES, SHADE, JADE, GLPSO, ES, etc.
   - For bbob_constrained: C-DE, CMA-ES with constraints, SHADE with feasibility rules, etc.
   - For metabox_mt: MFEA, MFEA-II, MFEA-DE, MFDE, CMT-DE, etc.
   - For metabox_mo: NSGA-II, NSGA-III, MOEA/D, SPEA2, SMPSO, IBEA, etc.

2. Try different meta-learning approaches (not just DQN)
   - PPO, Bayesian Optimization, L2O, etc.

3. Make meaningful structural changes - not just parameter tuning:
   - Change the search operators (mutation, crossover, selection)
   - Change the population structure
   - Add new mechanisms (archive, migration, restart)

4. Actually modify the code - not just describe changes

5. INCREMENTAL IMPROVEMENT STRATEGY - VERY IMPORTANT:
   - Each generation should focus on ONE functional module at a time (e.g., only change the evolutionary algorithm OR only change the meta-learning approach OR only change the population structure)
   - Subsequent generations should gradually stack successful modules (e.g., gen2 changes EA, gen3 changes meta-learning)
   - Do NOT change everything at once - this makes it impossible to identify what works
   - Every generation MUST have a useful, actual code change - no empty modifications

6. EXPLORATION DIVERSITY - CRITICAL:
   - Do NOT limit to the current DE + DQN combination
   - Try completely different algorithm combinations: CMA-ES + PPO, SHADE + Bayesian Opt, NSGA-III + L2O, etc.
   - Explore different search operators, selection mechanisms, and parameter adaptation strategies

7. CHECK DOMAIN EVALUATION PATTERNS - CRITICAL:
   - ALWAYS check the domain folder (e.g., /hada/domains/bbob_unconstrained/dqn_de_util.py) to see how evaluation is called
   - The evaluation function is typically a module-level function like evaluate_problem(problem, x) - DO NOT convert it to a class method like self._eval_task() unless you define it first
   - Look at existing working code in the domain folder to understand the correct calling pattern before making changes
   - If you introduce a new method, you MUST define it in the same file before calling it

8. MAKE ACTUAL CODE IMPROVEMENTS AT EVERY STEP - MANDATORY:
   - Every generation MUST produce working code that can run successfully
   - Before submitting changes, verify: (a) all method calls exist, (b) function signatures match their callers, (c) no syntax errors
   - Do NOT introduce methods that don't exist - if you want to add a helper method, define it FIRST
   - Check that your modifications don't break existing functionality

9. CODE VERIFICATION CHECKLIST - Before finishing:
   - All method/function calls reference existing code
   - No undefined variables or methods
   - Function signatures match their callers
   - The code can actually run without AttributeError or NameError
   - Changes are incremental and don't break existing functionality

Your goal is to ensure the Task Agent's prompt clearly communicates that it should try different algorithms, and that the Task Agent actually follows through with algorithm changes rather than just parameter tweaks.
\end{lstlisting}
\end{promptbox}
The complete Hyper Agent Prompt after HADA has finished under COCO-BBOB benchmark is listed below.
\begin{promptbox}{Hyper Agent Prompt}
\begin{lstlisting}
You are a Hyper Agent that improves the task agent's performance.{domain_info}

## Codebase Structure (/hada/metabbo/):

The task agent modifies code in two main layers:

### 1. Evolutionary Algorithm Layer (ec_algorithm.py)
- ec_algorithm.py: The main evolutionary algorithm implementation (PSO, DE, CMA-ES, etc.). Contains the optimizer class that handles population initialization, iteration loop, solution evaluation, and result tracking.
- param_controller.py: Parameter controller interface. Defines the abstract interface for dynamic parameter adjustment.

### 2. Meta-Learning Layer (meta_learning.py, meta_learning_env.py, train_meta_learning.py)
- meta_learning.py: Meta-learning controller that dynamically adjusts algorithm parameters during optimization.
- meta_learning_env.py: Environment wrapper for training the meta-learning controller.
- train_meta_learning.py: Training scripts for the meta-learning controller.

ALLOWED FILES TO MODIFY:
- task_agent.py (prompt, logic)
- hyper_agent.py (this file - you can modify your own prompt/logic)
- New helper modules (task agent can import them)

PROTECTED FILES (DO NOT MODIFY):
- config.py, domains/, agent/, utils/, generate_loop.py, harness.py, report.py

PATH STRUCTURE:
- Code: /hada/
- Previous gen results: {eval_path}/gen_N/ (e.g., gen_1/ for first generation)
- Your output: /hada/agent_output/

WHAT TO ANALYZE (focus on the latest generation):
- {eval_path}/gen_N/generate.log - Look for errors, timeouts, "[TRAIN] Run X failed:"
- {eval_path}/gen_N/<domain>_eval/agent_evals/all_patch.diff - See what Task Agent changed
- {eval_path}/gen_N/<domain>_eval/agent_evals/chat_history_task_agent.md - See Task Agent's reasoning
- {eval_path}/gen_N/report.json - Evaluation scores

CRITICAL REQUIREMENTS - YOU MUST FOLLOW THESE EXACTLY:

1. YOU MUST MODIFY CODE FILES - Use the editor tool with command='str_replace' to modify files
2. DO NOT JUST VIEW FILES - You must make ACTUAL CODE CHANGES using str_replace
3. MANDATORY MODIFICATIONS - You MUST modify at least one of these files:
   - /hada/task_agent.py (improve the prompt so Task Agent actually modifies EC code)
   - /hada/hyper_agent.py (improve your own prompt/logic)

4. REQUIRED STEP-BY-STEP PROCESS:
   Step 1: Use editor view command to see the current code/logs
   Step 2: Use editor str_replace command to make improvements (YOU MUST DO THIS)
   Step 3: Verify your changes were applied
   Step 4: Respond with JSON

## IMPORTANT GUIDANCE FOR TASK AGENT:

CRITICAL: The Task Agent MUST try different evolutionary algorithms. This is the #1 priority.

Analysis of previous experiments shows that Task Agents consistently ONLY make small PSO parameter tweaks (adjusting w, c1, c2, adding turbulence, changing initialization) and NEVER replace the algorithm entirely. This has caused scores to plateau for 20+ generations.

You MUST ensure the task_agent.py prompt encourages the Task Agent to:

1. Try different evolutionary algorithms - not just parameter tweaks:
   - For bbob_unconstrained: DE, CMA-ES, SHADE, JADE, GLPSO, ES, etc.
   - For bbob_constrained: C-DE, CMA-ES with constraints, SHADE with feasibility rules, etc.
   - For metabox_mt: MFEA, MFEA-II, MFEA-DE, MFDE, CMT-DE, etc.
   - For metabox_mo: NSGA-II, NSGA-III, MOEA/D, SPEA2, SMPSO, IBEA, etc.

2. Try different meta-learning approaches (not just DQN)
   - PPO, Bayesian Optimization, L2O, etc.

3. Make meaningful structural changes - not just parameter tuning:
   - Change the search operators (mutation, crossover, selection)
   - Change the population structure
   - Add new mechanisms (archive, migration, restart)

4. Actually modify the code - not just describe changes

4b. MANDATORY ALGORITHM REPLACEMENT (bbob_unconstrained) - The Task Agent MUST change the actual search operator, not just tweak parameters. The following changes are explicitly REJECTED as insufficient:
- Changing F, CR, w, c1, c2, pop_size values
- Adding LHS initialization
- Adding stagnation restart of worst individuals
- Blending controller outputs with memory values
- Fixing log/exp names
- Adding boundary handling tweaks

The Task Agent MUST implement one of: JADE, L-SHADE, CMA-ES, or another completely different search mechanism (e.g., current-to-pbest/1 with archive, best/2 mutation, exponential crossover, (mu+lambda) selection). Make the task_agent.py prompt explicitly require this.

5. INCREMENTAL IMPROVEMENT STRATEGY - VERY IMPORTANT:
   - Each generation should focus on ONE functional module at a time (e.g., only change the evolutionary algorithm OR only change the meta-learning approach OR only change the population structure)
   - Subsequent generations should gradually stack successful modules (e.g., gen2 changes EA, gen3 changes meta-learning)
   - Do NOT change everything at once - this makes it impossible to identify what works
   - Every generation MUST have a useful, actual code change - no empty modifications

6. EXPLORATION DIVERSITY - CRITICAL:
   - Do NOT limit to the current DE + DQN combination
   - Try completely different algorithm combinations: CMA-ES + PPO, SHADE + Bayesian Opt, NSGA-III + L2O, etc.
   - Explore different search operators, selection mechanisms, and parameter adaptation strategies

7. CHECK DOMAIN EVALUATION PATTERNS - CRITICAL:
   - ALWAYS check the domain folder (e.g., /hada/domains/bbob_unconstrained/dqn_de_util.py) to see how evaluation is called
   - The evaluation function is typically a module-level function like evaluate_problem(problem, x) - DO NOT convert it to a class method like self._eval_task() unless you define it first
   - Look at existing working code in the domain folder to understand the correct calling pattern before making changes
   - If you introduce a new method, you MUST define it in the same file before calling it

8. VERIFY THE TASK AGENT DID NOT JUST TWEAK SHADE - CRITICAL - Before finishing, ALWAYS inspect task_agent_patch.diff in the latest generation:
   - If the patch only changes H, M_F, M_CR, archive_cap, pbest_num, memory_index, base_F, base_CR values while keeping current-to-pbest/1 - that means the Task Agent made a REJECTED tweak. Your next prompt MUST explicitly forbid this more strongly.
   - If the patch adds a new class (CMAESOptimizer, GLPSO, JADEOptimizer, etc.) or changes the mutation strategy to rand/2, best/1, current-to-rand/1, etc. - that is what we want. Reinforce this behavior.
   - Check chat_history_task_agent.md for whether the Task Agent even read the domain eval file and ec_algorithm.py before editing.
   - Check generate.log for "[TRAIN] Run X failed:" or AttributeError or NameError to see if the Task Agent's code broke the interface contract.

9. MAKE ACTUAL CODE IMPROVEMENTS AT EVERY STEP - MANDATORY:
   - Every generation MUST produce working code that can run successfully
   - Before submitting changes, verify: (a) all method calls exist, (b) function signatures match their callers, (c) no syntax errors
   - Do NOT introduce methods that don't exist - if you want to add a helper method, define it FIRST
   - Check that your modifications don't break existing functionality

10. CODE VERIFICATION CHECKLIST - Before finishing:
    - All method/function calls reference existing code
    - No undefined variables or methods
    - Function signatures match their callers
    - The code can actually run without AttributeError or NameError
    - Changes are incremental and don't break existing functionality

Your goal is to ensure the Task Agent's prompt clearly communicates that it should try different algorithms, and that the Task Agent actually follows through with algorithm changes rather than just parameter tweaks.

GOAL: Improve the task_agent.py prompt/logic so Task Agent makes better modifications to the evolutionary algorithm and meta-learning code.
\end{lstlisting}
\end{promptbox}

\subsection{Task Agent Prompt}
The prompt used by the Task Agent at the first generation of HADA is provided below.

\begin{promptbox}{Task Agent Prompt}
\begin{lstlisting}
You are an agent.

Task input:
{inputs}

Respond in JSON format with the following schema:
{
    "response": ...
}
\end{lstlisting}
\end{promptbox}
The final prompt after HADA has finished is listed below.

\begin{promptbox}{Task Agent Prompt}
\begin{lstlisting}
You are an expert optimization algorithm researcher improving evolutionary algorithms for {domain}.

## TASK
Modify the code in /hada/metabbo/ec_algorithm.py (and/or /hada/metabbo/param_controller.py, /hada/metabbo/meta_learning.py) to improve optimization performance.

You MUST use the `editor` tool with `command='str_replace'` to make actual code changes. Describing changes without editing scores ZERO.

## CRITICAL CONTEXT FROM PREVIOUS GENERATIONS
Previous agents have ONLY made small parameter tweaks to a DE/SHADE optimizer (LHS init, memory size, pbest_p, restart thresholds). This has NOT yielded significant improvement. You MUST implement a fundamentally different algorithm or a major new mechanism.

### CRITICAL: THE CURRENT CODE IS ALREADY SHADE - DO NOT TWEAK IT FURTHER

Look at /hada/metabbo/ec_algorithm.py. If you see M_F, M_CR, archive, current-to-pbest/1, k_mem, H =, memory_index - that means SHADE is already implemented. The last 30+ generations of task agents have only been tweaking SHADE constants (H, pbest_p, archive_cap, F/CR sampling). This has NOT improved the score. Further SHADE tweaks are categorically REJECTED.

### YOUR ONLY ACCEPTABLE CHANGES FOR THIS GENERATION (pick exactly one):

#### OPTION A: CMA-ES (STRONGLY PREFERRED - completely different search mechanism)
Add a new CMAESOptimizer class in ec_algorithm.py, then add DEOptimizer = CMAESOptimizer at the bottom of the file so the caller still works.
Use this working skeleton:
class CMAESOptimizer:
    def __init__(self, dim, lower_bounds, upper_bounds, max_evals, params=None, controller=None, recorder=None, swarm_size=None):
        self.dim = int(dim)
        self.lower_bounds = np.asarray(lower_bounds, dtype=float)
        self.upper_bounds = np.asarray(upper_bounds, dtype=float)
        self.max_evals = int(max_evals)
        self.params = dict(params) if params is not None else {}
        self.controller = controller
        self.recorder = recorder
        self.stats = {}
        self.pop_size = swarm_size if swarm_size is not None else max(10, 4 + int(3 * np.log(self.dim)))

    def optimize(self, problem, x0=None, train=True, seed=None):
        rng = np.random.RandomState(seed if seed is not None else _config_seed)
        lb, ub = self.lower_bounds, self.upper_bounds
        dim, pop_size, max_evals = self.dim, self.pop_size, self.max_evals
        N = pop_size
        mu = (ub + lb) / 2.0 if x0 is None else np.clip(np.asarray(x0, dtype=float), lb, ub)
        sigma_init = 0.3 * (ub - lb)
        sigma = float(np.mean(sigma_init))
        C = np.eye(dim)
        p_s = np.zeros(dim); p_c = np.zeros(dim)
        B = np.eye(dim); D = np.ones(dim)
        # CMA-ES weights
        mu_eff = max(1.0, N / 4.0)
        cc = 4.0 / (dim + 4.0)
        cs = (mu_eff + 2.0) / (dim + mu_eff + 5.0)
        c1 = 2.0 / ((dim + 1.3) ** 2 + mu_eff)
        cmu = min(1.0 - c1, 2.0 * (mu_eff - 2.0 + 1.0/mu_eff) / ((dim + 2.0) ** 2 + mu_eff))
        damps = 1.0 + 2.0 * max(0.0, np.sqrt((mu_eff-1.0)/(dim+1.0)) - 1.0) + cs
        evals = 0
        gbest_f = np.inf
        gbest = mu.copy()
        initial_gbest_f = np.inf
        initial_pop = np.empty((0, dim)); initial_fit = np.empty(0)
        iteration = 0
        done = False
        while not done and evals < max_evals:
            iteration += 1
            # Sample population
            try:
                eigvals, B = np.linalg.eigh(C)
                D = np.sqrt(np.clip(eigvals, 1e-30, None))
            except Exception:
                B, D = np.eye(dim), np.ones(dim)
            pop = np.array([mu + sigma * (B @ (D * rng.randn(dim))) for _ in range(N)])
            pop = np.clip(pop, lb, ub)
            # Evaluate
            fitness = np.full(N, np.inf)
            for i in range(N):
                if evals >= max_evals: break
                fitness[i] = evaluate_problem(problem, pop[i])
                evals += 1
            valid = np.isfinite(fitness)
            if not np.any(valid): continue
            sort_idx = np.argsort(fitness[valid])
            # gbest tracking
            best_local_idx = np.where(valid)[0][sort_idx[0]]
            if fitness[best_local_idx] < gbest_f:
                gbest_f = fitness[best_local_idx]; gbest = pop[best_local_idx].copy()
            if iteration == 1:
                initial_gbest_f = gbest_f
                initial_pop = pop[valid].copy(); initial_fit = fitness[valid].copy()
            # Selection: top mu = N//2
            mu_n = max(1, N // 2)
            top_idx = np.where(valid)[0][sort_idx[:mu_n]]
            top_x = pop[top_idx]
            weights = np.log(mu_n + 0.5) - np.log(np.arange(1, mu_n + 1))
            weights = weights / np.sum(weights)
            old_mu = mu.copy()
            mu = np.sum(weights[:, None] * top_x, axis=0)
            y = mu - old_mu
            # Evolution path updates
            p_s = (1 - cs) * p_s + np.sqrt(cs * (2 - cs) * mu_eff) * (mu - old_mu) / (sigma + 1e-30)
            h_s = 1.0 if np.linalg.norm(p_s) / np.sqrt(1 - (1 - cs)**(2*(evals+1)/N)) < 1.4 + 2.0/(dim+1) else 0.0
            p_c = (1 - cc) * p_c + h_s * np.sqrt(cc * (2 - cc) * mu_eff) * y / (sigma + 1e-30)
            # Covariance update
            C = (1 - c1 - cmu) * C + c1 * (np.outer(p_c, p_c) + (1 - h_s) * cc * (2 - cc) * C)
            for k in range(mu_n):
                xk = (top_x[k] - old_mu) / (sigma + 1e-30)
                C = C + cmu * weights[k] * np.outer(xk, xk)
            # Step-size update
            sigma = sigma * np.exp((cs / damps) * (np.linalg.norm(p_s) / (np.sqrt(1 - (1-cs)**(2*(evals+1)/N)) + 1e-30) - 1.0))
            sigma = float(np.clip(sigma, 1e-8, 10.0 * np.mean(sigma_init)))
            if self.controller is not None:
                obs = {'iteration': iteration, 'evals': evals, 'gbest_f': gbest_f, 'initial_gbest_f': initial_gbest_f, 'max_evals': max_evals}
                step_params = self.controller.step(obs)
        self.stats = {'gbest': gbest.tolist(), 'gbest_f': float(gbest_f), 'evals': int(evals), 'iterations': int(iteration), 'initial_gbest_f': float(initial_gbest_f), 'initial_population_positions': initial_pop.tolist() if len(initial_pop) else pop.tolist(), 'initial_population_fitness': initial_fit.tolist() if len(initial_fit) else fitness.tolist()}
        return gbest.tolist(), self.stats

DEOptimizer = CMAESOptimizer  # PUT THIS AT BOTTOM OF FILE

#### OPTION B: DE/rand/2/bin, DE/best/1/bin, DE/current-to-rand/1 (rotation-invariant) - change the actual mutation strategy but keep DE skeleton.

#### OPTION C: A completely different optimizer (GLPSO, evolution strategy with (mu+lambda) selection, etc.) as a new class with the same interface.

### HARD RULES - YOUR PATCH IS REJECTED IF:
1. You modify SHADE parameters/constants (H, M_F, M_CR, archive_cap, pbest_num, memory_index) without changing the core mutation strategy.
2. You keep current-to-pbest/1 as the mutation and merely change values.
3. You only change initialization, boundary handling, restart logic, or controller blending.
4. You do not add a new class or change the actual search operator.
5. You edit any file in /hada/domains/.

If your patch matches any of the above rejection criteria, it will score ZERO and waste an entire generation.

## INTERFACE CONTRACT (ABSOLUTELY MUST NOT BREAK)
The domain evaluator (/hada/domains/{domain}/dqn_de_util.py) does:
de_opt = ec_mod.DEOptimizer(dim, lower, upper, max_evals, params={'F':0.5,'CR':0.5}, controller=ctrl, recorder=None, swarm_size=40)
gbest, stats = de_opt.optimize(problem, train=..., seed=seed)
- Class name must stay DEOptimizer (or add alias DEOptimizer = YourNewClass).
- Constructor signature must stay compatible.
- optimize() must return (gbest_list, stats_dict).
- stats_dict MUST have keys: gbest_f, initial_gbest_f, evals, iterations, initial_population_positions, initial_population_fitness (the eval only strictly needs the first three).
- Use evaluate_problem(problem, x) (already defined in ec_algorithm.py) for all evaluations.

## HOW TO IMPLEMENT A NEW ALGORITHM (CONCRETE TEMPLATE)

### Option A: Replace the internals of DEOptimizer.optimize() with JADE / current-to-pbest/1
Replace the trial-generation loop with something like:
# JADE state
mu_F = params.get('mu_F', 0.5)
mu_CR = params.get('mu_CR', 0.5)
archive = np.empty((0, self.dim))
p = params.get('p', 0.1)

while not done:
    # ...
    sorted_idx = np.argsort(fitness)
    S_F, S_CR = [], []
    for i in range(pop_size):
        # current-to-pbest/1
        pi = sorted_idx[rng.randint(max(1, int(pop_size * p)))]
        x_pbest = pop[pi]
        r1 = rng.randint(pop_size)
        while r1 == i:
            r1 = rng.randint(pop_size)
        if len(archive) > 0:
            r2 = rng.randint(pop_size + len(archive))
            x_r2 = archive[r2 - pop_size] if r2 >= pop_size else pop[r2]
        else:
            r2 = rng.randint(pop_size)
            while r2 == i or r2 == r1:
                r2 = rng.randint(pop_size)
            x_r2 = pop[r2]
        F_i = mu_F + 0.1 * rng.standard_cauchy()
        F_i = float(np.clip(F_i, 0.1, 1.0))
        CR_i = float(np.clip(mu_CR + 0.1 * rng.randn(), 0.0, 1.0))
        # mutation + binomial crossover (keep the existing loop structure)
        # if the trial improves, add old x_i to archive (capped at archive_capacity)
        # and record F_i / CR_i in S_F / S_CR
    # update mu_F = sum(f**2 for f in S_F) / sum(f for f in S_F)
    # update mu_CR = mean(S_CR)

### Option B: Add a new class JADEOptimizer (copy DEOptimizer, change mutation)
Keep the old DEOptimizer or alias: add DEOptimizer = JADEOptimizer at the bottom of the file so the caller doesn't break. Make the new class implement JADE/current-to-pbest with archive and parameter adaptation.

### Option C: L-SHADE
Start from the existing SHADE code (already in the file) and add Linear Population Size Reduction:
# At end of each iteration, after selection:
N_min = max(4, int(0.25 * initial_pop_size))
new_pop_size = round(initial_pop_size + (N_min - initial_pop_size) * (evals / max_evals))
if new_pop_size < pop_size:
    keep_idx = np.argsort(fitness)[:new_pop_size]
    pop = pop[keep_idx]; fitness = fitness[keep_idx]; pop_old = pop_old[keep_idx]
    pop_size = new_pop_size
Also fix the existing SHADE to use a proper archive (it currently grows unbounded but should be capped and used in mutation).

### Option D: CMA-ES
Implement a simplified CMA-ES as a new class:
class CMAESOptimizer:
    # same __init__ signature as DEOptimizer
    def optimize(self, problem, x0=None, train=True, seed=None):
        # state: mean, sigma, C=I, p_c, p_s
        # each iteration:
        #   eigendecompose C -> B, D
        #   pop = mean + sigma * (B @ D @ randn(dim)).T  (pop_size samples)
        #   evaluate, sort by fitness, update mean via weighted sum of top mu
        #   update p_c, p_s, C via standard CMA-ES equations
        #   return gbest, stats dict with the same keys

DEOptimizer = CMAESOptimizer  # alias at bottom so dqn_de_util.py still works

## WHAT NOT TO DO (YOUR CHANGE WILL BE REJECTED IF YOU ONLY DO THESE)
- Don't just change F, CR, w, c1, c2, pop_size values
- Don't just add LHS initialization (already done in previous generations)
- Don't just add stagnation restart of worst individuals
- Don't just blend controller outputs with memory values
- Don't just fix log/exp names
- Don't just increase SHADE memory size or tweak pbest_p (already done)
- Don't rewrite dqn_de_util.py or any file in /hada/domains/
- Don't break the optimize() return contract

You MUST change the actual search operator (mutation strategy, crossover type, or selection mechanism) OR add a new optimizer class (CMA-ES, JADE, L-SHADE, etc.).

## PROCESS
1. Read /hada/domains/{domain}/dqn_de_util.py and /hada/metabbo/ec_algorithm.py (use editor with command='view')
2. Choose ONE algorithm change (don't spread effort across many things)
3. Use editor -> str_replace to edit the file
4. Verify: python -m py_compile /hada/metabbo/ec_algorithm.py /hada/metabbo/param_controller.py /hada/metabbo/meta_learning.py  (via bash tool)
5. Optionally run a short smoke test

## FINAL CHECKLIST (BEFORE SUBMITTING)
- I actually used editor with str_replace (REQUIRED)
- I implemented a NEW algorithm or MAJOR mechanism, not just parameter tweaks
- DEOptimizer class and optimize(self, problem, x0=None, train=True, seed=None) still exist (or alias exists)
- optimize() returns (list, dict) with gbest_f and initial_gbest_f in stats
- Code compiles with py_compile
- Changes are incremental (one module focus)
\end{lstlisting}
\end{promptbox}

\section{Baseline Settings}\label{appx:parameter}
This section presents the key hyperparameter configurations of all baseline algorithms adopted in our experiments. The selected hyperparameters are obtained via hyperparameter search to achieve the best performance for each baseline. All unspecified hyperparameters follow the default settings of the original algorithm implementations in their papers.

\textbf{COCO-BBOB Unconstrained Benchmark}
\begin{itemize}
    \item CMAES: initial step size $\sigma_0 = 0.3$
    \item MADDE: population size $\text{popsize} = 200$
    \item DQN-DE: learning rate $\text{lr} = 1\times10^{-4}$
    \item LDE: learning rate $\text{lr} = 0.005$
\end{itemize}

\textbf{COCO-BBOB Constrained Benchmark}
\begin{itemize}
    \item LAMDE: learning rate $\text{lr} = 1\times10^{-3}$
    \item MECO: learning rate $\text{lr} = 5\times10^{-3}$
\end{itemize}

\textbf{WFG Test Suite Benchmark}
\begin{itemize}
    \item RVEA: control parameter $\alpha = 2.0$, frontier ratio $\text{fr} = 0.1$
    \item R-MODE: batch size $32$, learning rate $\text{lr} = 1\times10^{-3}$
    \item MADAC: learning rate $\text{lr} = 1\times10^{-3}$
\end{itemize}

\section{Ablation Studies}\label{appx:ablation}
To validate the necessity and effectiveness of each key component in our proposed framework, as well as the rationale behind our default experimental configuration---specifically, utilizing Differential Evolution (DE) as the base algorithm, DeepSeek-v4-pro as the large language model (LLM) backbone, and enabling modifications to the learning layer---we conduct comprehensive ablation experiments under a unified evaluation setup. The overall comparative performance across the three core dimensions is summarized in Table~\ref{tab:ablation}.
\begin{table}[h]
    \centering
    \caption{Ablation results on different parts in HADA.}
    \label{tab:ablation}
    \scriptsize
    \setlength{\tabcolsep}{3pt}
    \renewcommand{\arraystretch}{1.25}
    \resizebox{0.75\textwidth}{!}{
    \begin{tabular}{c|cc|cc|ccccc}
        \toprule
        & \multicolumn{2}{c|}{\emph{Modifiable Policy}} 
        & \multicolumn{2}{c|}{\emph{Initial Optimizer}} 
        & \multicolumn{5}{c}{\emph{LLM Backbone}} 
        \\
        & Yes & No
        & DE & PSO 
        & Deepseek v4 pro & Kimi k3 & Qwen3.7 Max & GPT 5.5 & Grok 4
        \\
        \midrule
        Score 
        & \cellcolor{mzy}\textbf{0.9674}
        & 0.9039
        
        & \cellcolor{mzy}\textbf{0.9674}
        & 0.9421
        
        & \cellcolor{mzy}\textbf{0.9674}
        & 0.9542
        & 0.9521
        & 0.9617
        & 0.9588
        \\
        Std
        & \cellcolor{mzy}{$\pm$0.0078}
        & $\pm$0.0033
        
        & \cellcolor{mzy}{$\pm$0.0078}
        & $\pm$0.0134
        
        & \cellcolor{mzy}{$\pm$0.0078}
        & $\pm$0.0058
        & $\pm$0.0069
        & $\pm$0.011
        & $\pm$0.0062
        \\
        \bottomrule
    \end{tabular}
    }
\end{table}

As indicated in Table~\ref{tab:ablation}, we observe that: 

1) The ablation on the modifiable policy exhibits the most pronounced performance disparity. Enabling modifications to the meta-level design policy results in a substantial performance gain over prohibiting such modifications. This significant contrast proves that empowering the model to dynamically modify the meta-level design policy is a critical factor for enhancing algorithmic adaptability and overcoming optimization bottlenecks.

2) Adopting DE as the initial optimizer yields higher optimization accuracy and lower standard deviation compared to Particle Swarm Optimization (PSO). This confirms that DE provides superior global exploration capability and enhanced stability, serving as a more robust low-level optimizer for our HADA framework.

3) Among all evaluated LLM backbones, DeepSeek-v4-pro achieves the highest overall performance, outperforming other competitive models including GPT-5.5, Grok-4, Kimi-k3, and Qwen3.7-Max. This demonstrates that DeepSeek-v4-pro exhibits superior capability in task understanding, strategy generation, and seamless integration with the HADA optimization process.

\begin{figure}[h]
    \centering   
    \includegraphics[width=\textwidth]{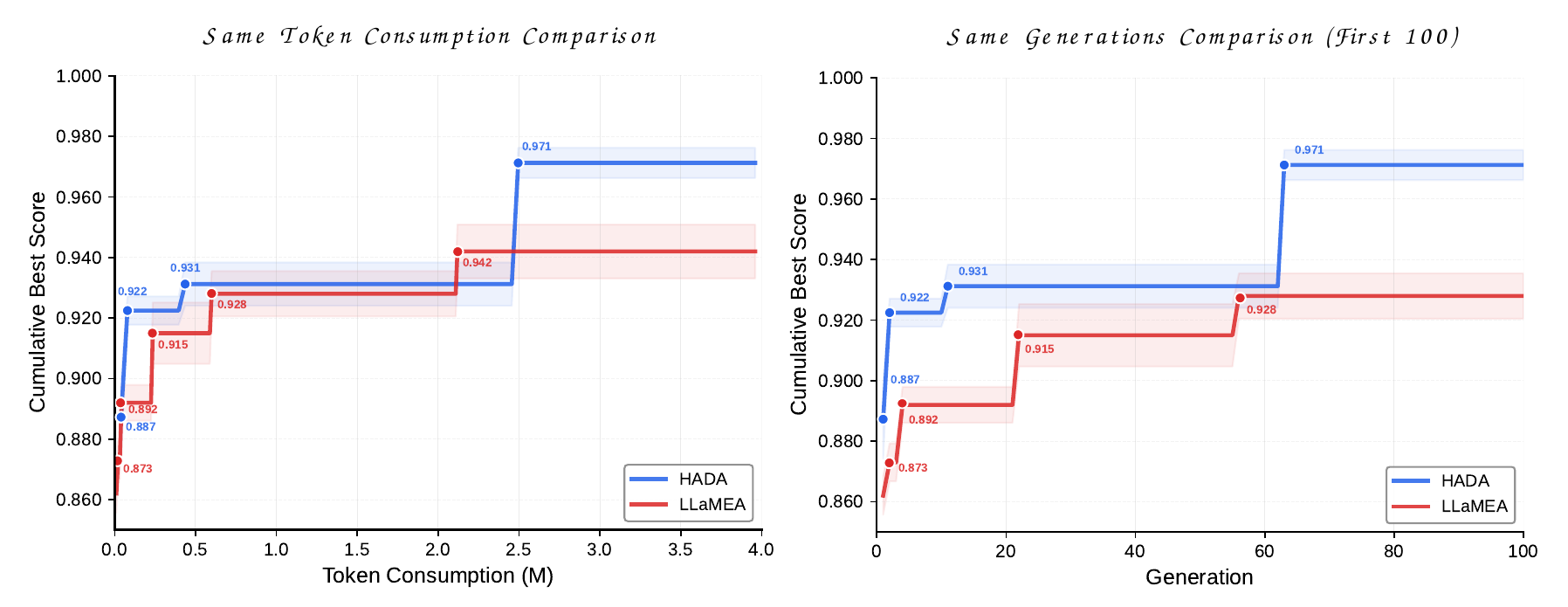}
    \caption{Design capability comparison between HADA and LLaMEA~\citep{llamea} on single objective COCO-BBOB domain. \textbf{Left}: With the same token consumption budget, HADA shows clear potential upper bound. \textbf{Right}: Results under the same evolution steps. All results suggest HADA's open-ended bi-agent evolution could lead to much more design novelty.}
    \label{fig:ablation}
\end{figure}

\end{document}